\documentclass[sigconf]{acmart}

\AtBeginDocument{%
  \providecommand\BibTeX{{%
    \normalfont B\kern-0.5em{\scshape i\kern-0.25em b}\kern-0.8em\TeX}}}

\setcopyright{acmcopyright}
\copyrightyear{2020} 
\acmYear{2020} 
\setcopyright{acmcopyright}\acmConference[KDD '20]{Proceedings of the 26th ACM SIGKDD Conference on Knowledge Discovery and Data Mining}{August 23--27, 2020}{Virtual Event, CA, USA}
\acmBooktitle{Proceedings of the 26th ACM SIGKDD Conference on Knowledge Discovery and Data Mining (KDD '20), August 23--27, 2020, Virtual Event, CA, USA}
\acmPrice{15.00}
\acmDOI{10.1145/3394486.3403221}
\acmISBN{978-1-4503-7998-4/20/08}

\usepackage[utf8]{inputenc}
\usepackage{amsmath}
\usepackage{tabularx}
\usepackage{graphicx}
\usepackage{color}
\usepackage{multirow}
\usepackage{enumitem}
\usepackage{xcolor}

\begin{document}
\fancyhead{}
\title{Interpretable Deep Graph Generation with Node-edge Co-disentanglement}

\author{Xiaojie Guo}
\email{xguo7@gmu.edu}
\affiliation{%
  \institution{George Mason University}
  \city{Fairfax}
  \state{VA}
  \country{USA}
}

\author{Liang Zhao}
\authornote{Corresponding author: lzhao9@gmu.edu}
\email{lzhao9@gmu.edu}
\affiliation{%
  \institution{George Mason University}
  \city{Fairfax}
  \state{VA}
    \country{USA}
}

\author{Zhao Qin}
\email{zqin02@syr.edu}
\affiliation{%
  \institution{Syracuse University}
  \city{Syracuse}
  \state{NY}
    \country{USA}
}

\author{Lingfei Wu}
\email{wuli@us.ibm.com}
\affiliation{%
  \institution{IBM Research AI}
  \city{Yorktown Heights}
  \state{NY}
  \country{USA}
}

\author{Amarda Shehu}
\email{ashehu@gmu.edu}
\affiliation{%
  \institution{George Mason University}
  \city{Fairfax}
  \state{VA}
    \country{USA}
}

\author{Yanfang Ye}
\email{yanfang.ye@case.edu}
\affiliation{%
  \institution{Case Western Reserve University}
  \city{Cleveland}
  \state{OH}
    \country{USA}
}

\begin{abstract}
   {\color{black}Disentangled representation learning has recently attracted significant amount of attentions, particularly in the field of image representation learning. However, learning the disentangled representations behind a graph remains largely unexplored, especially for the attributed graph with both node and edge features. Disentanglement learning for graph generation has substantial new challenges including: 1) the lack of graph deconvolution operations to jointly decode node and edge attributes; and 2) the difficulty in enforcing the disentanglement among latent factors that respectively influence: i) only nodes, ii) only edges, and iii) joint patterns between them.
   To address these challenges, we propose a new disentanglement enhancement framework for deep generative models for attributed graphs. In particular, a novel variational objective is proposed to disentangle the above three types of latent factors, with novel architecture for node and edge deconvolutions. Moreover, within each type, individual-factor-wise disentanglement is further enhanced, which is shown to be a generalization of existing framework for images. Qualitative and quantitative experiments on both synthetic and real-world datasets demonstrate the effectiveness of the proposed model and its extensions.}
\end{abstract}


\maketitle

\section{Introduction}
Recent advances in deep generative models, such as variational auto-encoders (VAE)~\cite{kingma2013auto} and generative adversarial networks (GAN)~\cite{goodfellow2014generative}, have made important progress towards generative modeling for complex domains, such as image data. The goal here is to learn the underlying (low-dimensional) distribution of the images, hence image generation is treated as sampling from learned distributions. Building on these techniques for images, which can be considered as grid-structured data, a special case of graphs, a number of deep learning models for generating general graphs have been proposed over the last couple of years~\cite{li2018learning,kipf2016semi,simonovsky2018graphvae}. These involves real-world applications such as modeling physical and social interactions~\cite{kusner2017grammar,dai2018syntax}, discovering new chemical and molecular structures, and constructing knowledge graphs.

\begin{figure}[htb]
    \centering
    \includegraphics[width=0.45\textwidth]{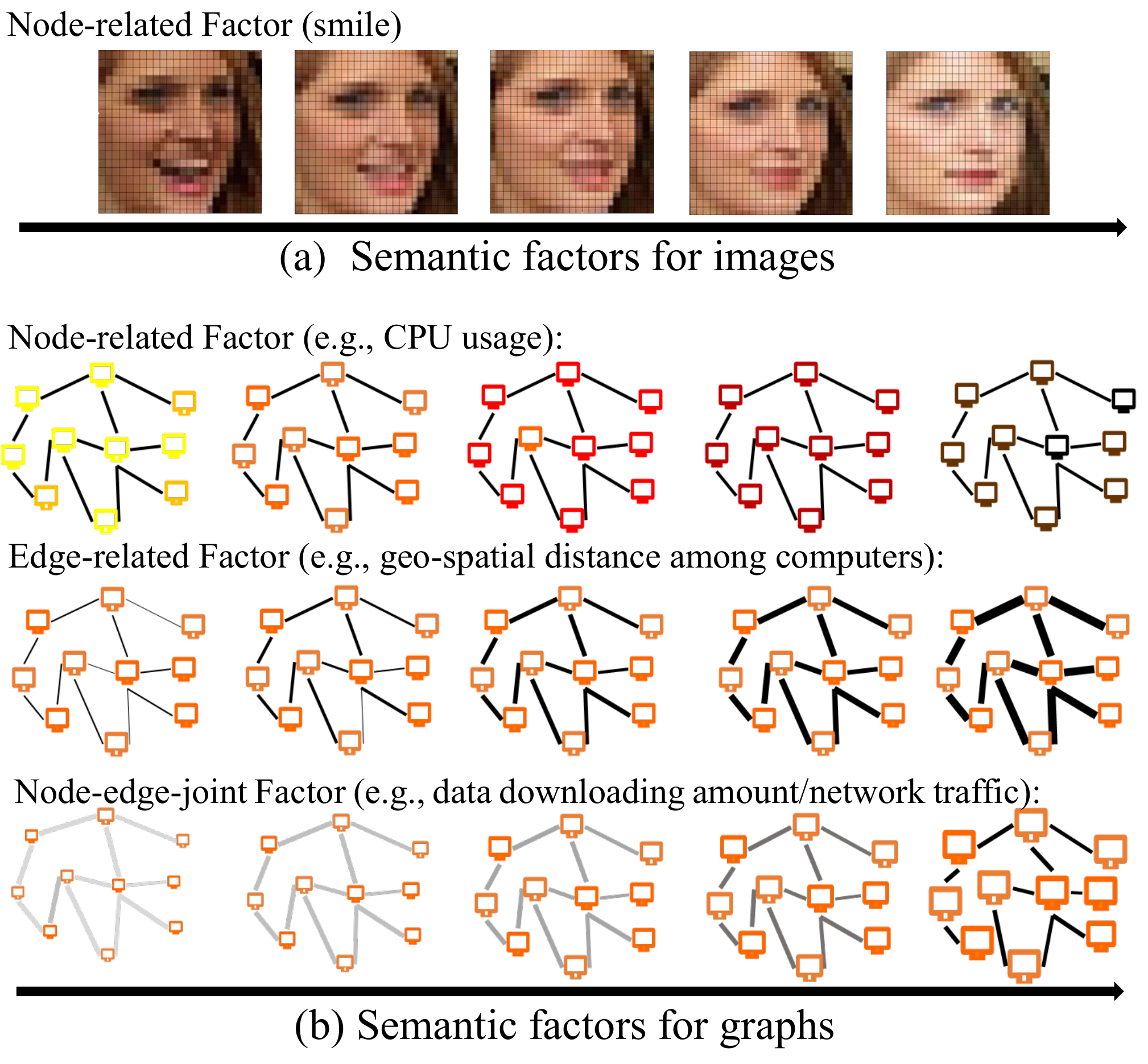}
    \caption{Two examples of disentanglement: (a) semantic factors of images, where each pixel is a node and each pixel is connected to its eight neighboring pixels, and (b) semantic factors of cyber networks, where each computer is a node and the link between each pair of computers is an edge (better seen in color).}
    \label{fig:example}
\end{figure}

When we learn the underlying distribution of complex data such as images, learning interpretable representations of data that expose semantic meaning is very important. Such representations are useful not only for standard downstream tasks such as supervised learning and reinforcement learning, but also for tasks such as transfer learning and zero-shot learning where humans excel but machines struggle~\cite{lake2017building}. As yet, most research has focused on learning factors of variations in the data, commonly referred to as learning a disentangled representation, where the variables of the representation are highly independent. Examples of this include variables that only control the size of objects, or their color. For the instance in Fig.~\ref{fig:example} (a), where a semantic factor controls the degree of smile in a human facial image. 

However, in the promising domain of deep generative models for graph generation, disentangled enhancement has rarely been well explored yet, but could be highly beneficial for applications such as controlling the generation of protein structures, or designing Internet of Things (IoT). As shown in Fig.~\ref{fig:example}, we would love to generalize from an image situation to a graph situation, where the variables control specific factors related to node attributes, edge attributes, or joint-node-edge patterns in the graph. For example, Fig.~\ref{fig:example} (a) shows the semantic factor (i.e. smile) in the images, which can be regarded as special cases of graphs where nodes are pixels that are connected in a fixed topology. All the factors that control image formulation are effectively node-related. Fig.~\ref{fig:example}(b) shows the factors that control the formulation of a cyber network, which is an attributed graph where computers are nodes and their links are edges. Unlike images, there are three types of factors formulating the networks: (1) node-related factors that control some properties of node attributes but are independent of edge patterns (e.g., the CPU usage of each computer); (2) edge-related factors that only influence edge patterns but are independent of node patterns (e.g., geo-spatial distances between computers); and (3) node-edge-joint factors that jointly influence some properties from both nodes and edges (e.g., the node patterns of "downloaded data amount" and edge patterns "network traffic" which are inherently highly entangled and hence must be controlled by such factors). Thus, it is necessary to develop a generic model to discover and disentangle all three types of factors for the graph data. Though a few researchers have sought to apply the disentanglement learning to graphs~\cite{ma2019learning,liu2019independence,stoehr2019disentangling}, as yet they only have identified the latent factors that caused the edge between a node and its neighbors. 

In this paper, we focus on the generic problem of disentanglement learning on attributed graphs, where the characteristics of graphs pose great challenges to disentanglement learning on graphs:~1)~\textbf{Lack of node and edge joint deconvolution operations}. The formation process for real-world graphs, which is both complex and iterative, is based on the three types of factors depicted in Fig.~\ref{fig:example}. For example, edges are generated not only by the edge related factors, but also node-edge-joint related factors. There is no existing graph decoder that can simultaneously handle all three types of factors during the generation process.
2)~\textbf{Complex disentanglement enhancement of multiple types of latent representation}. Although the three types of semantic factors mentioned in Fig.~\ref{fig:example} are independent from each other, it is extremely difficult to enforce that. First, it is difficult to automatically categorize individual factors into these three types. Second, even they are categorized, the enhancement of such independency patterns still cannot be accomplished by the existing techniques which mostly focus on images without categorization capability. 
3)~\textbf{The dilemma between
disentanglement and reconstruction quality for attributed graphs}. Disentangling the three types of factors and reconstructing both edges and nodes can require multiple trade-offs between reconstruction errors and disentanglement performance during the training process. For example, the objective of disentanglement of node-edge-joint factors can conflict with not only edge but also node reconstruction errors. Existing methods cannot handle the situation of attributed graphs.

To the best of our knowledge, this is the first work that can address all the above challenges and provides a generic
framework that incorporates multiple disentanglement enhancements for attributed graphs. We propose the new Node-Edge Disentangled Variational Auto-encoder (NED-VAE) model, a deep unsupervised generative approach for disentanglement learning on graphs that automatically discovers the independent latent
factors in both edges and nodes. A novel objective for node-edge jointly disentanglement is derived and proposed based on the variational autoencoder (VAE)~\cite{kingma2013auto,rezende2014stochastic}. A novel architecture is proposed consisting of three sub-encoders and two sub-decoders to model the complicated relationships between nodes and edges. 
We also propose a general framework of objectives that can include various extensions of the base NED-VAE to realize the group-wise and variable-wise disentanglement. The
contributions of this work are summarized as follows:

\begin{itemize} \vspace{-0.1cm}
    \item\textbf{A novel framework is proposed for the disentanglement of attributed graph generation}. In order to jointly disentangle the nodes and edges, we derive a novel objective framework for learning three factors that are exclusive to node patterns, exclusive to edge patterns, and those spanning node-edge-joint patterns. 
    This new framework is demonstrated to be a significant generalization over existing disentanglement frameworks for image generation.
    \item\textbf{A novel architecture is proposed for disentanglement learning on graphs.} Derived from the theoretical objective of our framework, a novel architecture proposed for the representation learning of graphs consists of three sub-encoders (a node encoder, an edge encoder, and a node-edge co-encoder) to learn the three types of representations, along with two novel sub-decoders (a node-decoder and an edge decoder) to co-generate both nodes and edges.
    \item\textbf{Simultaneous group-wise and variable-wise disentanglement}. The proposed framework hierarchically disentangles attributed graph generation according to node, edge, and their joint factors. A set of varational auto-encoder-based models for attributed graphs have been proposed. 
    \item\textbf{Comprehensive experiments have been conducted to validate the effectiveness of our proposed model and its extensions}. Qualitative and quantitative experiments on two synthetic and two real-world datasets demonstrate that NED-VAE and its extensive models are indeed capable of learning disentangled factors for different types of graphs.
\end{itemize}

\section{Related Works}
\textbf{Disentanglement Learning}.
Disentangled representation learning has gained considerable attention, in particular in the field of image representation learning~\cite{higgins2017beta,alemi2016deep,chen2018isolating,kim2018disentangling}. The goal is to learn representations that separate out the underlying explanatory factors responsible for variations in the data. Such representations have been shown to be relatively resilient to the complex variants involved~\cite{bengio2013representation}, and can be used to enhance generalizability as well as improve robustness against adversarial attack~\cite{alemi2016deep}. The disentangled representations are inherently more interpretable, and can thus potentially facilitate debugging and auditing~\cite{doshi2017towards}.
This has prompted a number of approaches that modify the VAE objective by adding, removing, or altering the weight of individual terms~\cite{kim2018disentangling,chen2018isolating,zhao2017infovae,kumar2017variational,lopez2018information,esmaeili2019structured,alemi2016deep}. However, the best way of learning representations that disentangle the latent factors behind a graph remains largely unexplored.

\textbf{Graph neural networks}.
Recent work on graph neural networks (GNNs)~\cite{gori2005new,scarselli2008graph}, especially graph convolutional networks~\cite{bruna2013spectral,henaff2015deep}, is
attracting considerable attention, because of their remarkable success in multiple domains such as natural language processing ~\cite{chen2019reinforcement,chen2019graphflow}, computer vision~\cite{Shen2020hierarchical}, software engineering~\cite{leclair2020improved} and traffic flow prediction~\cite{li2019large}. Graph Convolutional Networks originated from spectral graph convolutional neural networks~\cite{bruna2013spectral}, which were then extended by using fast localized convolutions~\cite{defferrard2016convolutional}, and further approximated
by an efficient architecture for a semi-supervised setting proposed by~\citet{kipf2016semi}. Self-attention mechanisms and sub graph-level information have also been explored as ways to potentially improve the representation power of learned node embeddings~\cite{velivckovic2017graph,bai2019unsupervised,gao2019dyngraph2seq}. 

\textbf{Graph generation}. Most of the existing GNN based graph generation methods are based on VAE~\cite{simonovsky2018graphvae,samanta2018designing} and generative adversarial nets (GANs)~\cite{bojchevski2018netgan}, and others~\cite{li2018learning, you2018graphrnn}. For example, GraphRNN~\cite{you2018graphrnn} builds an autoregressive generative model on these sequences utilizing LSTM model and has demonstrated good scalability; while GraphVAE~\cite{simonovsky2018graphvae} represents each graph in terms of its adjacent matrix and feature vector and utilizes the VAE model to learn the distribution of the graphs conditioned on a latent representation at the graph level. Graphite~\cite{grover2019graphite} and VGAE~\cite{kipf2016variational} encode the nodes of each graph into node-level embeddings and predict the links between each pair of nodes to generate a graph. Some conditional graph generation methods also provide powerful graph encoders and decoders for attributed graphs where both node and edge attributes are considered~\cite{guo2018deep,guo2019deep}. 

\section{Problem Formulation}
Define an input graph as $G(\mathcal{V},\mathcal{E},E,F)$, where $\mathcal{V}$ is the set of $N$ nodes and $\mathcal{E}\subseteq \mathcal{V} \times \mathcal{V}$ is the set of $M$ edges. 
$\mathcal{E}$ contains all pairs of nodes, while the existence of each edge is reflected by one of its attributes. $E\in \mathbb R^{ N\times N\times K}$ is the edge attributes tensor, 
where $K$ is the dimension of the edge attributes. $F\in \mathbb R^{ N\times D}$ refers to the node attribute matrix, where $F_{i}\in \mathbb R^{1\times D}$ is the node attributes of node $i$ and $D$ is the dimension of the node attribute vector. As shown in Fig.~\ref{fig:example}, three types of factors (i.e. node-related factors, edge-related factors and node-edge-joint related factors) are assumed to control the generation of the graph $G$.

The goal is to develop an unsupervised deep generative model that can learn the joint distribution of the graph $G$ and three groups of generative latent variables $Z=(z_e\in \mathbb{R}^{L_1},z_f\in \mathbb{R}^{L_2},z_g\in \mathbb{R}^{L_3})$ ( $L_1$, $L_2$, and $L_3$ are the number of variables in each group) to discover the three types of factors, such that the observed graph $G$ can be generated as $p(G|Z)=P(E,F|z_e,z_f,z_g)$. There are three challenges must be overcome to achieve the above goal: (1) The lack of co-decoder based on co-deconvolution for the generation of attributed graph $G$ that is capable of jointly generating both the nodes attributes $F$ and edges attributes $E$; (2) difficulty of enforcing independence among the variable groups $z_e$, $z_f$ and $z_g$ (group-wise disentanglement), rather than simply enforcing the disentanglement of the variables inside $z_e$, $z_f$ and $z_g$ (variable-wise disentanglement); and (3) the need to simultaneously solve multiple reconstruction-disentanglement conflicts in $z_e$ and $E$,  $z_f$ and $F$, $z_g$ and $E$, and $z_g$ and $F$.

\section{Node-edge Disentanglement VAE}
In this Section, we first introduce the derived training objective and the architecture of the proposed Node-edge Disentanglement VAE (NED-VAE).
Then we propose a generic objective framework as well as its derivation to further enforce the disentanglement of NED-VAE models with different purposes. At last, the time and memory complexities of the proposed NED-VAE are analyzed and compared with the existing methods.

\subsection{Objective and Architecture}
In this section, we first derive the objective for learning disentanglement on graphs. Then, to solve the first challenge, we propose a new architecture, the NED-VAE, based on the derived objectives. NED-VAE includes a novel co-deconvolution-based co-decoder that is capable of jointly generating nodes and edges.  
\subsubsection{The objective for disentanglement on graphs}
Inspired by the disentanglement learning in the image domain, a suitable objective is to maximize the marginal (log-)likelihood of the observed graph $G$ in expectation over the whole distribution of latent factors set Z:

\begin{equation}
  \max_{\theta} \mathbb{E}_{p_{\theta}(Z)} [p_{\theta}(E,F|z_e,z_f,z_g)].
\end{equation}

For a given observation $G=(E,F)$, we describe the inferred posterior configurations of the latent factors $Z=(z_e,z_f,z_g)$ using a probability distribution $q_{\phi}(z_e,z_f,z_g|E,F)$. Our aim is to ensure that the inferred latent factors $q_{\phi}(z_e,z_f,z_g|E,F)$ capture all three types of generative factors in a disentangled manner. In order
to encourage this disentangling property in the inferred $q_{\phi}(z_e,z_f,z_g|E,F)$, we can introduce a constraint by trying to match it to a prior $p(z_e)$, $p(z_f)$ and $p(z_g)$ that both controls the capacity of the latent information bottleneck,
and embodies the statistical independence mentioned above. This can be achieved if we set the prior to be an isotropic unit Gaussian, i.e.~$p(Z)=p(z_e,z_f,z_g)=\mathcal{N}(\textit{0},\textit{1})$, leading to the constrained optimisation problem in Eq.~\ref{infer1}, where $\epsilon$ specifies the strength of the applied constraint:
\begin{align}\nonumber
    \max_{\theta,\phi} &\mathbb{E}_{G\thicksim D}[\mathbb{E}_{q_{\phi}(Z|G)}log p_{\theta}(E,F|z_e,z_f,z_g)]   \\
    &s.t. D_{KL}(q_{\phi}(z_e,z_f,z_g|E,F)||p(z_e,z_f,z_g)<\epsilon.
    \label{infer1}
\end{align}

Eq.~\ref{infer1} can be rewritten as a Lagrangian under the KKT conditions and, according to the complementary slackness KKT condition, we therefore arrive at the $\beta$-VAE~\cite{higgins2017beta} formulation, which takes the form of the familiar variational free energy objective function:
\begin{align}\nonumber
    \mathcal{L}(\theta,\phi,G,Z,\beta)=& \mathbb{E}_{q_{\phi}(Z|G)}[log p_{\theta}(E,F|z_e,z_f,z_g)]\\
    &-\beta D_{KL}(q_{\phi}(z_e,z_f,z_g|E,F)||p(z_e,z_f,z_g).
    \label{infer2}
\end{align}

Based on the definitions of $z_f$, $z_e$, and $z_g$, namely that $z_f$ only controls some properties of nodes, $z_e$ only controls some properties of edges and $z_g$ controls the properties of both, we obtain:
\begin{align}
    q_{\phi}(z_e,z_f,z_g|E,F)=q_{\phi}(z_f|F)q_{\phi}(z_e|E)q_{\phi}(z_g|E,F)\\
    p_{\theta}(E,F|z_e,z_f,z_g) =p_{\theta}(F|z_f,z_g)p_{\theta}(E|z_e,z_g)
    \label{eq:bayes}
\end{align}

We can now rewrite the loss function as:
\begin{align}\nonumber
   \mathcal{L}(\theta,\phi,&G,Z,\beta)
    =\mathbb{E}_{q_{\phi}(Z|G)}[logp_{\theta}(F|z_f,z_g)p_{\theta}(E|z_e,z_g)]\\\nonumber
    &-\beta D_{KL}(q_{\phi}(z_f|F)q_{\phi}(z_e|E)q_{\phi}(z_g|E,F)||p(z_e)p(z_f)p(z_g))\\\nonumber
    &=\mathbb{E}_{q_{\phi}(Z|G)}[logp_{\theta}(F|z_f,z_g)p_{\theta}(E|z_e,z_g)]\\\nonumber
    &-\beta D_{KL}(q_{\phi}(z_f|F)||p(z_f))\\\nonumber
    &-\beta D_{KL}(q_{\phi}(z_e|E)||p(z_e))\\
    &-\beta D_{KL}(q_{\phi}(z_g|E,F)||p(z_g))
\end{align}
Given that the goal is to maximize the above objective, a deep generative model is needed to model each of the components in this objective. 

\subsubsection{The architecture of the node-edge disentangled VAE}
\begin{figure}[htb]
    \centering
    \includegraphics[width=0.45\textwidth]{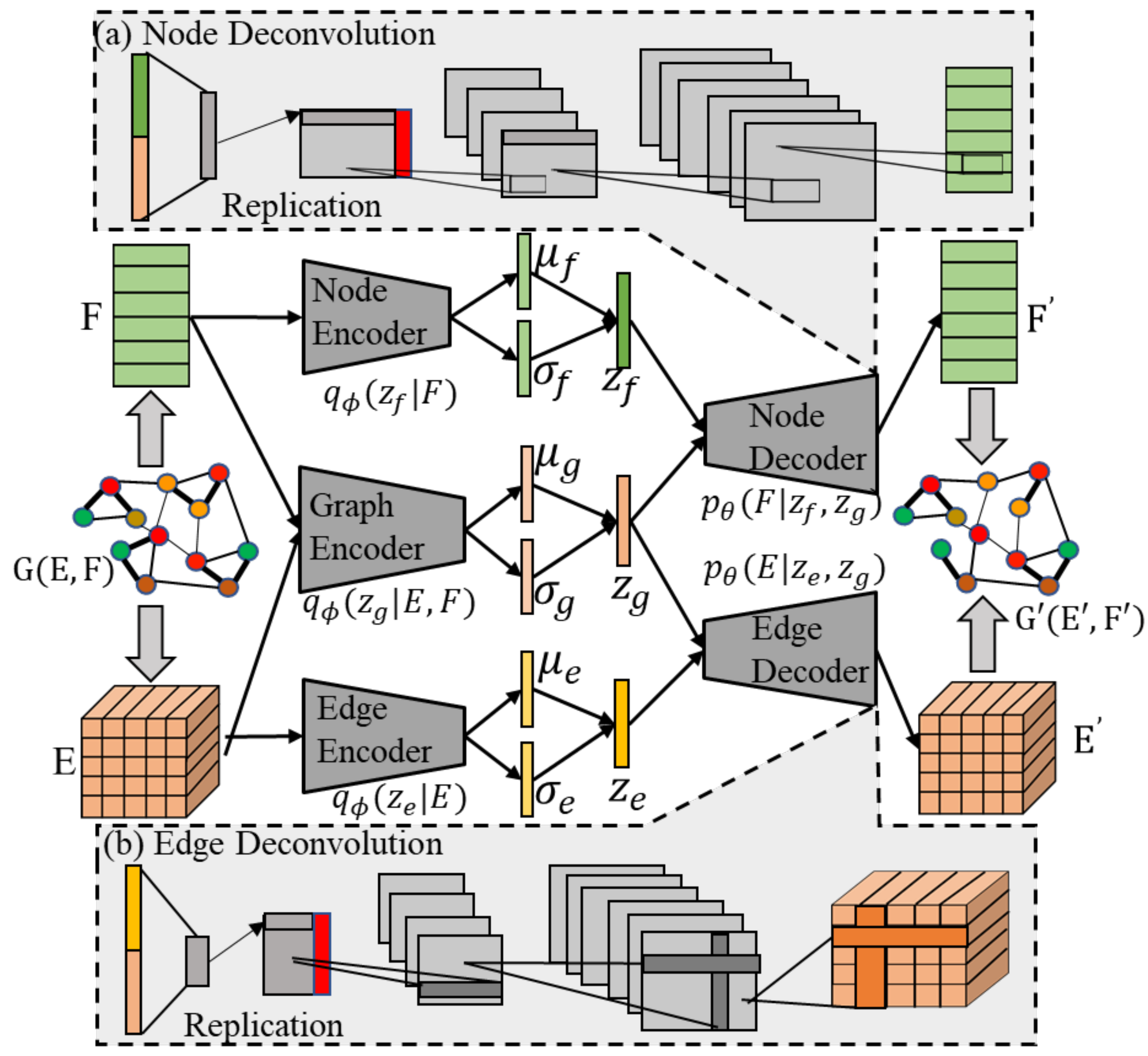}
    \caption{The architecture of the proposed NED-VAE consist of three sub-encoders to inference $z_e$, $z_f$ and $z_g$, as well as two sub-decoders to reconstruct $E$ and $F$ simultaneously.}
    \label{fig:architecture}
\end{figure}

Based on the above inference for the objective, we are proposing the Node-Edge Disentangled VAE model (NED-VAE) based on a novel architecture. The architecture of the proposed model is shown in Fig.~\ref{fig:architecture}. 

The overall framework is based on the traditional VAE, where the encoder learns the mean and standard deviation of the latent representation of the input and the decoder decodes the sampled latent representation vector to reconstruct the input. Unlike the structure of traditional VAE, the proposed framework has three encoders, each of which models the distributions $q_{\phi}(z_f|F)$, $q_{\phi}(z_g|E,F)$ or $q_{\phi}(z_e|E)$; and two novel decoders to model $p_{\theta}(F|z_g,z_f)$ and $p_{\theta}(E|z_g,z_e)$, that jointly generate the node and edge attributes based on the three types of latent representations. Each type of representations is sampled by their own inferenced mean and standard derivation. For example, the representation vectors $z_f$ are sampled as $z_e=\mu_f+\sigma_f\odot\epsilon$, where $\epsilon$ follows a standard normal distribution. This architecture also partially solves the second challenge described above because it enforces the disentanglement between the two groups of variables $z_e$ and $z_f$ by separating their inference process. The details of each components are described as follows.

\paragraph{\textbf{Node, edge and graph encoder}.} 
The \textit{node encoder} consists of several traditional convolution layers to extract latent features from node attribute matrix $F$; and two paths of fully connected layers to get the mean $\mu_f$ and standard derivation vectors $\sigma_f$ of the node representation distribution.
The \textit{edge encoder} consists of several edge convolution layers proposed by \citet{guo2018deep} to extract edge representations from the edge attribute tensor $E$; and edge embedding layers to get the node-level representation;
and fully connected layers to yield the mean $\mu_e$ and standard derivation $\sigma_e$ vectors of the edge representation distribution.
The \textit{graph encoder} consists of several graph convolution layers proposed in~\cite{kipf2016semi} to get node-level representations; and fully connected layers to aggregate the learned node representations into a graph-level representation that can be separately mapped into the mean $\mu_g$ and standard derivation $\sigma_g$ vectors of the graph representation distribution~\footnote{Operation details of the encoders can be found in \url{https://github.com/xguo7/NED-VAE}.}.

\paragraph{\textbf{Node decoder}.}
The proposed node decoder aims to generate the node attribute matrix $F'$ based on the sampled node representations $z_f$ and graph representations $z_g$, which ensures the node attribute generation process is controlled by both the node-related factors and the node-edge-joint related factors.
As shown in Fig.~\ref{fig:architecture} (a), the node decoder consists of several traditional deconvolution layers and fully connected layers as a reversed process of the node encoder. First, the input node representation $z_f$ and graph representation $z_g$ are concatenated together and mapped into several fully connected layers to decode the vector into multiple feature vectors. Next, we aim to convert each feature vector into a feature matrix, where each row should refer to an individual node, by replicating each feature vectors $N$ times. Moreover, to ensure the diversity and randomness of the nodes in each graph, a node assignment vector $S\in\mathbb{R}^{N}$ (shown as a red rectangle in Fig.~\ref{fig:architecture} (a) is sampled following the normal distribution and is concatenated with each feature matrix. Thirdly, once the feature matrix has been obtained, one-dimensional filters are used to deconvolute each row of the feature matrix into the attribute vectors for each node, completing the reconstruction of the input node attribute matrix $F$.

\paragraph{\textbf{Edge Decoder}.}
The proposed edge decoder aims to generate the reconstructed edge attribute $E$ based on the sampled node representations $z_e$ and graph representations $z_g$, ensuring that the edge attributes generation is controlled by both the edge-related and node-edge-joint related factors. The proposed edge decoder consists of several edge deconvolution layers and fully connected layers as a reversed process of the edge encoder. The input is the concatenation of both the edge representation $z_e$ and the graph representation $z_g$. First, the input vector is mapped into a node-level feature vector through a fully connected layer and is converted into a matrix by being replicated. The same node assignment vector $S$ is also concatenated to this feature matrix. The hidden edge feature matrices are then generated by the edge-node deconvolution layer~\citep{guo2018deep} by decoding each of the node-level representations, where the principle is that each node's representation can make contributions to the generation of its related edges features (contributions are shown as dark grey rectangles in Fig.~\ref{fig:architecture} (b)). Thirdly, the edge-attribute tensor $E$ is generated through the edge-edge deconvolution layer, where the principle is that each hidden edge feature can contribute to the generation of its adjacent edges.

\subsection{Framework of node-edge co-disentanglement} 
To solve the second and third challenges, we propose a generic objective framework to further enforce the disentanglement of NED-VAE models with different purposes. In Section~\ref{sec:framework}, the basic overall framework with four terms are introduced, namely two conditional distribution terms of the graphs (denoted as $\textcircled{1}$), the latent representations term (denoted as $\textcircled{2}$), the marginal distribution term for the graphs (denoted as $\textcircled{3}$), and the inferred prior distributions (denoted as $\textcircled{4}$). In Section \ref{sec:inferred priors}, we move on to further enforce the disentanglement among variable groups as explained in the second challenge, generalizing the term $\textcircled{4}$ to introduce a novel node-edge-total-correlation term (denoted as $\textcircled{A}$) for group-wise disentanglement and a variable-wise disentanglement term (denoted as $\textcircled{C}$). Next, in Section \ref{sec:variable-wise}, we further enforce the disentanglement inside the three types of latent representations, generalizing the term $\textcircled{C}$ to introduce three variable-total-correlation terms (denoted as $\textcircled{A}_f$, $\textcircled{B}_f$, and $\textcircled{C}_f$).   
Furthermore, based on the framework, six extensions of the base NED-VAE models are proposed that enforce different terms, as shown in Table ~\ref{tab:disen-graph-VAE-models}. The existing disentanglement methods in the image domain are proven to be special cases of our generic framework.

\subsubsection{Overall graph disentanglement framework}
\label{sec:framework}
As proved by~\citet{esmaeili2019structured}, the VAE objective can be equivalently defined as a KL divergence between the generative model $p_{\theta}(x,z)$ and inference model $q_{\phi}(z,x)=q_{\phi}(z|x)q(x)$. Inspired by this and given that $\small{p(z_1,z_2,z_3)=p(z_1)p(z_2)p(z_3)}$, in conjunction with Eq.~\ref{eq:bayes}, the NED-VAE objective for the graph data can be defined as:
\begin{align}\nonumber\tiny 
    &\mathcal{L}(\theta,\phi,G,Z,\beta)
    =-D_{KL}(p_{\theta}(z_e,z_f,z_g,E,F)||q_{\phi}(E,F,z_e,z_f,z_g))\\\nonumber
    &=\mathbb{E}_{q_{\phi}(Z,G)}[log\frac{p_{\theta}(E,F,z_e,z_f,z_g)}{p_{\theta}(E,F)p(z_e,z_f,z_g)}+log\frac{q(E,F)q_{\phi}(z_e,z_f,z_g)}{q_{\phi}(E,F,z_e,z_f,z_g)}\\\nonumber
    &+log\frac{p_{\theta}(E,F)}{q(E,F)}+log\frac{p(z_e,z_f,z_g)}{q_{\phi}(z_e,z_f,z_g)}]\\\nonumber
    &=\mathbb{E}_{q_{\phi}(Z,G)}[log\frac{p_{\theta}(E,F|z_e,z_f,z_g)}{p_{\theta}(E,F)}-log\frac{q_{\phi}(z_e,z_f,z_g|E,F)}{q_{\phi}(z_e,z_f,z_g)}]\\\nonumber
    &-KL(q(E,F)||p_{\theta}(E,F))-KL(q_{\phi}(z_e,z_f,z_g)||p(z_e,z_f,z_g))\\\nonumber
     &=\mathbb{E}_{q_{\phi}(Z,G)}[\underbrace{log\frac{p_{\theta}(F|z_f,z_g)p_{\theta}(E|z_e,z_g)}{p_{\theta}(E,F)}}_{\textcircled{1}}\\\nonumber
     &-\underbrace{log\frac{q_{\phi}(z_e|E)q_{\phi}(z_f|F)q_{\phi}(z_g|E,F)}{q_{\phi}(z_e)q_{\phi}z_f)q_{\phi}(z_g)}}_{\textcircled{2}}]\\
    &\underbrace{-KL(q(E,F)||p_{\theta}(E,F)))}_{\textcircled{3}}-\underbrace{KL(q_{\phi}(z_e,z_f,z_g)||p(z_f)p(z_e)p(z_g))}_{\textcircled{4}}
    \label{eq: decompose}
\end{align}



Specifically, Terms \textcircled{3} and \textcircled{4} enforce consistency between the marginal distributions over $G=(E,F)$ and $Z=(z_e,z_f,z_g)$. Minimizing the KL divergence in Term \textcircled{3} maximizes the marginal likelihood $\mathbb{E}_{q(E,F)}logp_{\theta}(E,F)$; maximizing Term \textcircled{4} which is named as inferred priors term enforces the distance between $q_{\phi}(z_e,z_f,z_g)$ and $p(z_e,z_f,z_g)$. Terms \textcircled{1} and \textcircled{2} enforce consistency between the conditional distributions. Specifically, Term \textcircled{1} maximizes the correlation for each $Z$ that generates each $G^n$; when $Z\sim q_{\phi}(Z|G^n)$ is sampled, the likelihood $p_{\theta}(G^n|Z)$ should be higher than the marginal likelihood $p_{\theta}(G^n)$. Meanwhile Term \textcircled{2} regularizes Term \textcircled{1} by minimizing the mutual information $I(Z,G)$ in the inference model.

Since Term \textcircled{2} actually represents the mutual information between the latent $z_e$, $z_f$, $z_g$ and the graphs $G$, this will lead to poor reconstructions when enforcing disentanglement with high values of $\beta$ in the proposed NED-VAE~\cite{makhzani2017pixelgan}. Thus, to solve the trade-off problems between the disentanglement of $z_e$, $z_f$, $z_g$ and $G$, we propose to either enforce Term \textcircled{4} alone or enforce it with high weights. Accordingly, we can refer to the model enforcing only Term \textcircled{4} as (Node-edge disentangled Inferred Priors VAE) NED-IPVAE-I, and the model enforcing both \textcircled{2} and \textcircled{4} with different weights as NED-IPVAE-II, as shown in Table~\ref{tab:disen-graph-VAE-models}. 
\begin{table}[htb]
   \caption{Summary of objectives of the extensions of NED-VAE model. ($\textcircled{\small{C}}^*$ refers to the sum of $\textcircled{\small{C}}_e$, $\textcircled{\small{C}}_f$ and $\textcircled{\small{C}}_g$; $\textcircled{2}^a_e$ can be changed to $\textcircled{2}^a_f$ or $\textcircled{2}^a_g$)}
    \centering
    \begin{tabular}{l|c}
    \hline
         NED-VAE& \textcircled{1}+\textcircled{3}+$\beta$(\textcircled{2}+\textcircled{4}) \\\hline
        NED-IPVAE-I & \textcircled{1}+\textcircled{3}+\textcircled{2}+$\lambda$\textcircled{4}\\\hline
        NED-IPVAE-II&\textcircled{1}+\textcircled{3}+$\lambda$\textcircled{4}\\\hline
        NED-HCVAE&\textcircled{1}+\textcircled{3}+\textcircled{2}+$\gamma$\textcircled{A}\\\hline
        NED-TCVAE& \textcircled{1}+\textcircled{3}+\textcircled{2}+\textcircled{\small{C}}+$\beta$\textcircled{A}\\\hline
        NED-VTCVAE&\small\textcircled{1}+\textcircled{3}+\textcircled{2}+\textcircled{\small{C}}*+$\beta$\textcircled{A}+$\gamma_1\textcircled{A}_f$+$\gamma_2\textcircled{A}_e$+$\gamma_3\textcircled{A}_g$\\\hline
        NED-AnchorVAE&\textcircled{1}+\textcircled{3}+\textcircled{2}+\textcircled{4}-$\lambda\textcircled{2}^a_e$\\\hline
    \end{tabular}
    \label{tab:disen-graph-VAE-models}
\end{table}

\subsubsection{Generalization of the Inferred Priors Term \textcircled{4}}
\label{sec:inferred priors}
Next, to further address the second challenge and enforce the disentanglement among groups of variables $z_e$, $z_f$ and $z_g$, we further generalize the Term \textcircled{4} by decomposing it and introduce the Node-edge Total Correlation term (\textcircled{A} in Table.~\ref{tab:disen-graph-VAE-models}). Specifically, Term \textcircled{4} can be decomposed into sub components \textcircled{A}, \textcircled{B} and \textcircled{C}, as the followings (Here, we use $Z$ to denote $(z_e,z_f,z_g)$ for clarity):
\begin{align}\nonumber\scriptsize
    &-D_{KL}(q_{\phi}(Z)||p(z_e)p(z_f)p(z_g)\\\nonumber &=-\mathbb{E}_{q_{\phi}(Z)}[log\frac{q_{\phi}(Z)}{q_{\phi}(z_e)q_{\phi}(z_f)q_{\phi}(z_g)}+log\frac{q_{\phi}(z_e)q_{\phi}(z_f)q_{\phi}(z_g)}{p(z_e)p(z_f)p(z_g)}\\\nonumber
    &+log\frac{p(z_e)p(z_f)p(z_g)}{p(Z)}]\\ \nonumber
    &=-E_{q_{\phi}(z)}\underbrace{[log\frac{q_{\phi}(Z)}{q_{\phi}(z_e)q_{\phi}(z_f)q_{\phi}(z_g)}}_{\textcircled{A}}+\underbrace{log\frac{p(z_e)p(z_f)p(z_g)}{p(Z)}]}_{\textcircled{B}}\\\nonumber
    &-\underbrace{D_{KL}(q_{\phi}(z_e)||p(z_e))-D_{KL}(q_{\phi}(z_f)||p(z_f))-D_{KL}(q_{\phi}(z_g)||p(z_g))}_{\textcircled{C}}
\end{align}

We refer to Term \textcircled{A} as the ``Node-Edge Total Correlation'' term since it measures the dependence between the three types of latent of graphs $z_e$, $z_f$ and $z_g$ (group-wise disentanglement). The penalty for this term forces the model to find statistically independent factors for the nodes, the edges and their combinations. A heavier penalty on this term induces better separately and disentangled learning for the graph format data. We refer to Term \textcircled{C} as the ``variable-disentangelment'' term which enforces the disentanglement of the variables inside each latent group. This allows us to propose variant model which only penalizes Terms \textcircled{A} and \textcircled{C}, shown as the Node-edge Disentangled Total Correlation VAE (NED-TCVAE) in Table~\ref{tab:disen-graph-VAE-models}. In some application cases where only the group-wise disentanglement is needed, and the variable-wise disentanglement in $z_e$, $z_f$ and $z_f$ is not required. This kind of disentanglement can be referred to as a ``Half Correlation Disentanglement" of the graphs, where the penalty for Term \textcircled{C} is ignored, leading to another variant model NED-HCVAE, as defined in Table.\ref{tab:disen-graph-VAE-models}.

When calculating Term \textcircled{A}, we utilize the Na\"ive Monte Carlo approximation based on a mini-batch of samples to underestimate $q_{\phi}(Z)$, $q_{\phi}(z_e)$, $q_{\phi}(z_f)$, and $q_{\phi}(z_g)$, as described in work proposed by~\citet{chen2018isolating}.


\subsubsection{Generalization of variable-wise disentanglement \textcircled{C}}
\label{sec:variable-wise}
To further enforce the variable-wise disentanglement, we generalize Term \textcircled{C} by decomposing it to obtain the ``Variable Total Correlation“(VTC) terms to largely enforce the variable-wise disentanglement in $z_e$, $z_f$ and $z_g$ respectively. The following shows the decomposition of $D_{KL}(q_{\phi}(z_f)||p(z_f))$ in Term \textcircled{C} as an example:
\begin{align}\tiny\nonumber
    &-D_{KL}(q_{\phi}(z_f)||p(z_f))\\\nonumber
    &=-\mathbb{E}_{q_{\phi}(z_f)}[log\frac{q_{\phi}(z_f)}{\prod_d q_{\phi}(z_f^d)}+log\frac{\prod_d q_{\phi}(z_f^d)}{\prod_d p(z_f^d)}+log\frac{\prod_d p(z_f^d)}{p(z_f)}]\\\nonumber
    &=-\mathbb{E}_{q_{\phi}(z_f)}\underbrace{[log\frac{q_{\phi}(z_f)}{\prod_d q_{\phi}(z_f^d)}}_{\textcircled{A}_f}-\underbrace{log\frac{p(z_f)}{\prod_d p(z_f^d)}]}_{\textcircled{B}_f}
    -\sum_d \underbrace{D_{KL}(q_{\phi}(z_f^d)||p(z_f^d))}_{\textcircled{C}_f}
\end{align}
Here, Term $\textcircled{A}_f$ (referred to as the ``Node Total Correlation“(TC)) is the most important term as it helps the model to identify the statistically independent factors in the representation $z_f$, as proved by~\citet{watanabe1960information}. Similarly, when decomposing the latent $z_e$ and $z_g$, we obtain their respective TC terms $\textcircled{A}_e$ and $\textcircled{A}_g$. The relevant variant model, labelled $NED-VTCVAE$ in Table.~\ref{tab:disen-graph-VAE-models}, can flexibly enforce both the group-wise disentanglement and the variable-wise disentanglement with pre-defined weights.

\subsubsection{Generalization of conditional distribution Term \textcircled{2}}
\label{sec:mutual term}
In some cases, we are really only concerned with node attributes or edge attributes, so we need only control either the nodes or edges when generating the graph. Thus, to learn the types of factors involved, we can anchor a single group of latent variable (e.g., $z_e$), to yield higher mutual information with the observation graphs $G$.

First, if we decompose Term \textcircled{2} in Eq.~\ref{eq: decompose}, we have:
\begin{align}\nonumber\tiny
    &-log\frac{q_{\phi}(z_e|E)q_{\phi}(z_f|F)q_{\phi}(z_g|E,F)}{q_{\phi}(z_e)q_{\phi}(z_f)q_{\phi}(z_g)}\\
    &=\underbrace{log\frac{q_{\phi}(z_e)}{q_{\phi}(z_e|E)}}_{\textcircled{2}^a_e}+\underbrace{log\frac{q_{\phi}(z_f)}{q_{\phi}(z_f|F)}}_{\textcircled{2}^a_f}+\underbrace{log\frac{q_{\phi}(z_g)}{q_{\phi}(z_g|E,F)}}_{\textcircled{2}^a_g}.
    \end{align}

Since each of the three above terms actually represents mutual information between observations and latent representations, because $-log\frac{q_{\phi}(z_e)}{q_{\phi}(z_e|E)}=-log\frac{q_{\phi}(z_e)q_{\phi}(E)}{q_{\phi}(z_e,E)}=log\frac{q_{\phi}(z_e,E)}{q_{\phi}(z_e)q_{\phi}(E)}=I(z_e,E)$. Thus, enforcing them can help ensure the mutual information between each types of latent representations and observed graphs. The extensive model that enforces either of the three terms is named as NED-AnchorVAE in Table~\ref{tab:disen-graph-VAE-models}.

\subsubsection{Relation to existing models}
Next, we demonstrate that the existing disentanglement methods, where only the disentanglement representation learning of nodes attributes is considered, are actually special cases of our proposed new frameworks.

First, as a special case of attributed graph, image only involves node attributes and node-related factors matters. Hence in this special case, the NED-VAE objective can be rewritten by ignoring $z_e$ and $z_g$ as: 
\begin{align}\nonumber
   \mathcal{L}(\theta,\phi,G,Z,\beta)&
    =\mathbb{E}_{q_{\phi}(Z|F)}[logp_{\theta}(F|z_f)
    -\beta D_{KL}(q_{\phi}(z_f|F)||p(z_f)),
\end{align}
which is the same objective as that defined in $\beta-VAE$~\cite{higgins2017beta} for the image domain. In the same way, we can easily demonstrated that $DIP-VAE$~\cite{kumar2017variational} is a special case of the the proposed $NED-IPVAE-I$, obtained by enforcing the inferred priors disentanglement, and $InforVAE$ is a special case of the proposed $NED-IPVAE-II$.  

In addition, the proposed NED-TCVAE is a more general form that includes the objective of two existing methods $FactorVAE$~\cite{kim2018disentangling} and $\beta-TCVAE$~\cite{chen2018isolating} which share the same objectives. For example, when the weight $\beta$ of Term \textcircled{A} is $0$, and there is no $z_e$ and $z_g$, there is no need to enforce the group-wise disentanglement among the edge-latent $z_e$, node-related latent $z_f$ and node-edge joint latent $z_g$. Only the variable-wise disentanglement $\textcircled{C}_f$ is used.

\subsection{Complexity Analysis}
The proposed NED-VAE requires $O(N^2)$ operations in time complexity and $O(N^2)$ computation complexity in terms of number of nodes in the graph, which paves the way toward modest scale graphs with hundreds or thousands of nodes, compared to most of the existing graph generation methods, which often have $O(N^3)$ or even $O(N^4)$ computational costs. For example, GraphVAE~\cite{simonovsky2018graphvae} requires $O(N^4)$ operations in the worst case and \citet{li2018learning} uses graph neural networks to perform a form of message passing with $O(MN^2)$ operations to generate a graph.

\section{Experiment}
This section reports the results of both qualitative and quantitative experiments that are carried out to test the performance of NED-VAE and its extensions on two synthetic and one real-world datasets. All experiments are conducted on a 64-bit machine with an NVIDIA GPU (GTX 1070, 1683 MHz, 16 GB GDDR5)~\footnote{The code of the model and additional experiment results and details are available
at:~\url{https://github.com/xguo7/NED-VAE}.}.

\subsection{Dataset}
\subsubsection{Erdos-Renyi Graphs}
Erdos-Renyi (ER) graphs are generated based on three types of factor. One is an edge-related factor $a$ that refers to the probability of edge creation in a graph following the rule specified in \cite{erdHos1960evolution}; the second is a node-related factor $b$ which is the mean of a Gaussian random distribution (the standard is set to 0.1), based on which node attribute $F_{i,1}$ is generated; and the third is a node-edge-joint related factor $c$ defining the function: $F_{i,2}=degree(i)+10*c$ (where $c$ is a positive integers chosen from 1 to 10), based on which the second node attribute $F_{i,2}$ is generated. Here, $degree(i)$ refers to the degree of Node $i$. The dimension of the node attribute and edge attribute is 2 and 1 respectively. A total of 25,000 ER graphs are used for training and 12,500 for testing.

\subsubsection{Watts Strogatz Graphs}
Watts Strogatz (WR) graphs are also generated based on three types of factor. One is an edge-related factor $a$ that indicates the number of nearest neighbours that each node is joined to in a ring topology~\cite{watts1998collective}; the second is a node-related factor $b$ that refers to the mean of a Gaussian distribution (the standard is set as 0.01) based on which node attribute is generated; and the third factor is a node-edge-joint related factor $c$ that not only defines the probability of rewiring each edge for graph topology but also defines the second node attribute as:$F_{i,2}=degree(i)+10*c$. The dimension of the node attribute and edge attribute is 2 and 1 respectively. A total 25,000 WR graphs used for training and 12,500 for testing.

\subsubsection{Protein Structure Dataset}
Protein structures can be formulated as graph structured data where each amino acid is a node and the geo-spatial distances between them are edges. To generate the dataset, we simulate the dynamic folding process of a protein peptide with a sequence AGAAAAGA, which for our purposes can be considered as a graph of 8 nodes with node attributes $(x,y,z)$ corresponding to 3D coordination of the $C_\alpha$ atom of each amino acid. The protein contact map (graph topology) is generated based on fully atomistic molecular dynamics simulations. There are two factors involved in generating the contact maps and nodes attributes: simulation time (T) and ionic concentration (C), both of which are edge-related factors. Here, 38 values are used for the ionic concentration (C) and 2,000 values are used for the simulation time (T) to generate the dataset, producing 38,000 samples for training and 38,000 samples for testing.  

\subsection{Comparison Methods}
Since graphVAE~\cite{simonovsky2018graphvae} is the only existing method that fits the requirement of graph disentanglement (i.e, not only learning the representations of graphs but also generate both edge and node attributes), it is utilized as one comparison method. In addition, to validate the necessities of inferring three types of representations separately, a baseline model called GDVAE is used, which has only one graph encoder for inferring an overall graph representation vector. The proposed model NED-VAE as well as the extensions (except NED-AnchorVAE) in Table~\ref{tab:disen-graph-VAE-models} are all tested and compared.  

\subsection{Evaluation Metrics}
\subsubsection{Qualitative Metrics}
As it is important to be able to measure the level of disentanglement achieved by different models, we search to qualitatively demonstrate that our proposed NED-VAE model and its extensions consistently discover more latent factors and disentangles them in a cleaner fashion than the previous models. By learning a latent code representation of a graph, we assume that each variable in the latent code corresponds to a certain factor or property that is used to generate the graphs' edge and node attributes. Thus, by changing the value of one variable continuously and fixing the remaining variables, we can visualize the corresponding change in the generated graphs.

\subsubsection{Quantitative Metrics}
We used four quantitative metrics to evaluate the disentanglement of the proposed models. $\beta-M$~\cite{higgins2017beta} measures disentanglement by examining the accuracy of a linear classifier that predicts the index of a fixed factor of variation; while
$F-M$~\cite{kim2018disentangling} addresses several issues by using a majority vote classifier on a different feature vector that represents a corner case in the $\beta-M$; 
and the modularity score (mod)~\cite{ridgeway2018learning} measures whether each dimension of $z$ depends on at most a factor describing the maximum variation using their mutual information. Finally, disentanglement metric, DCI metric~\cite{eastwood2018framework} computes the entropy of the distribution obtained by normalizing the importance of each dimension of the learned representation for predicting the value of a factor of variation. 
All the implementation details are the same as those in the work proposed by~\citet{locatello2019challenging}.

\subsection{Results for ER dataset}

\subsubsection{Qualitative Evaluation}
\begin{figure*}[htb]
    \centering
    \includegraphics[width=0.95\textwidth]{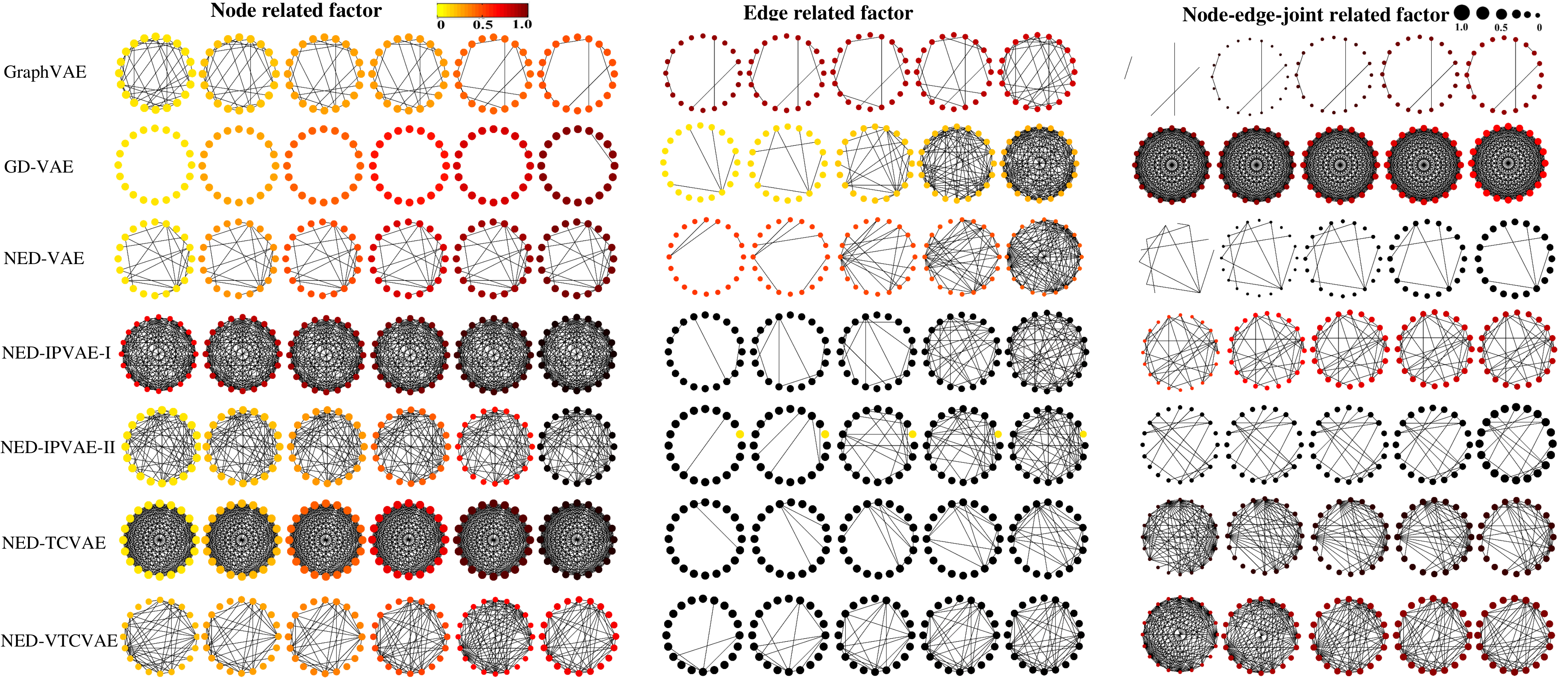}
    \caption{Generated Graphs from different models when the related latent variables range from 0 to 10 for ER graphs: (a) node-related factor which is reflected by color of node icon; (b) edge-related factor which is reflected by edge density (c) node-edge-joint related factor which is reflected by the size of node icon.}
    \label{fig:traveling_z_1}
\end{figure*}

For ER graphs visualization, the color of nodes is used to represent the value of the node-related factor $b$, and graph topology is used to represent the value of the edge-related factor $a$, and the size of the node is used to represent the value of the edge-node-combined factor $c$. The values of the latent variables range in $[0,10]$ and some segments of the generated graphs is shown in Fig.~\ref{fig:traveling_z_1}. All of the proposed node-edge disentangelment models (NED-) shows the best capabilities in discovering and disentangling all the three types of factors than the graphVAE and the baseline GVAE. For example, the node-related factor travels well with the obvious color ranging, while the discovered node-related factor by graphVAE is not disentangled well because it has some influence on the edges. This is highly due to the powerful co-decoder in the generation of both nodes.

\begin{table}[htb]\small
\renewcommand\arraystretch{0.9}
    \caption{Comparison of disentanglement scores of the proposed NED-VAE and its extensions for three datasets.}
    \centering
    \begin{tabular}{l|l|rrrrrrrrr}
    \hline\hline
    Dataset&method &$\beta$-M(\%)&F-M(\%)&DCI&Mod\\\hline
     \multirow{5}{*}{ER}
     &GraphVAE&79.20 &33.30 &0.33&0.75\\
     ~&GDVAE&79.20&33.34&0.33&0.74 \\
    ~&NED-VAE&97.20 &86.70&0.62&0.95 \\
    ~& NED-IPVAE-I &99.71 &\textbf{98.84}&\textbf{0.73}&0.92 \\
    ~& NED-IPVAE-II&\textbf{99.90}&98.70&0.71&0.93\\
    ~& NED-TCVAE&99.70&88.00&0.64&0.92 \\
    ~& NED-VTCVAE&94.00&59.10&0.63&\textbf{0.97} \\\hline
     \multirow{5}{*}{WS}
    &GraphVAE&73.10 &37.87&0.13&0.49\\
    ~&GDVAE&73.06&37.86&0.13&\textbf{0.62} \\
    ~&NED-VAE&\textbf{100.00}&64.96&0.16&0.52\\
    ~&NED-IPVAE-I &99.30&91.23&0.16&0.50 \\
    ~&NED-IPVAE-II&\textbf{100.00}&\textbf{97.82}&0.16&0.50 \\
    ~&NED-TCVAE&94.91&64.70&0.16&0.50 \\
    ~&NED-VTCVAE&94.50&49.33&\textbf{0.17}&0.51\\\hline
     \multirow{5}{*}{Protein}
    &GraphVAE&54.00 &50.00&0.20&0.61\\
    ~&GDVAE&54.00&50.00&0.21&0.60 \\
    ~&NED-VAE&\textbf{63.42}&61.67&\textbf{0.31}&\textbf{0.69} \\
    ~&NED-IPVAE-I &60.46&55.20&\textbf{0.31}&0.67 \\
    ~&NED-IPVAE-II&60.00&\textbf{64.00}&0.28&0.67 \\
    ~&NED-TCVAE&57.63&50.25&0.25&0.68 \\
    ~&NED-VTCVAE&58.40&50.00&0.24&0.67\\\hline
    \hline
    \end{tabular}
    \label{tab:all_quantitative}
\end{table}

\subsubsection{Quantitative Evaluation}
Four quantitative evaluation metrics are tested on different models and compared in Table~\ref{tab:all_quantitative}. The proposed node-edge disentanglement models all shows greater superority than graphVAE and baseline GDVAE. Specifically, NED-IPVAE-II achieves the score of $99.90\%$ in $\beta-M$, outperforming comparison methods by $20\%$ and other proposed extensions by $2.5\%$. NED-IPVAE-I achieves $98.84\%$ score in $F-M$, outperforming comparison methods by $66.28\%$ and other proposed extensions by $16.9\%$.The great superiority of the two NED-IP-VAE models is mainly due to their great penalty on the inferred prior term in the objective, which balances the trade-off between the reconstruction error and the disentanglement.


\subsection{Results for WR dataset}
\subsubsection{Qualitative Evaluation}
\begin{figure*}[htb]
    \centering
    \includegraphics[width=0.95\textwidth]{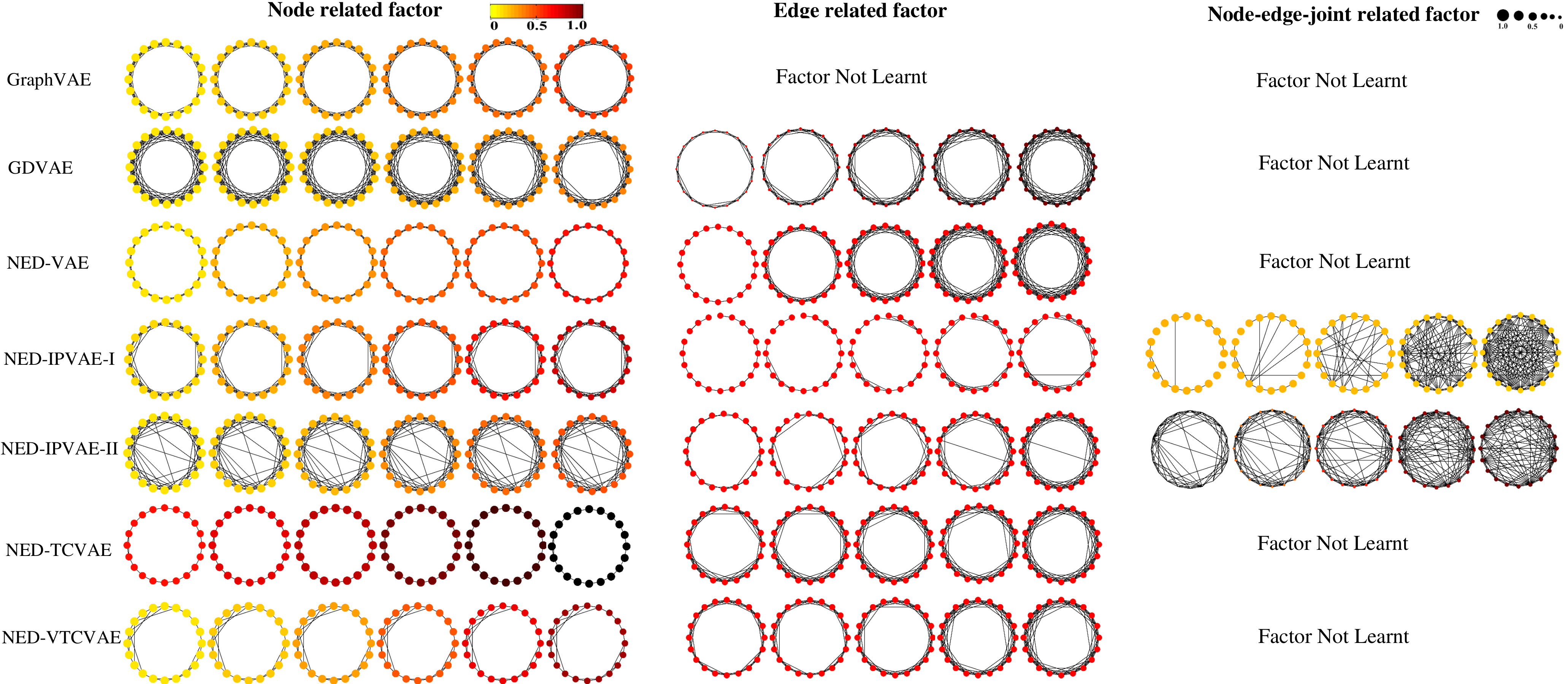}
    \caption{Generated Graphs from different graph disentangled models when the related latent variable value ranges from 0 to 10 for WS graphs: (a) node-related factor, which is reflected by color of node icon; (b) edge-related factor which is reflected by number of rings in topology and (c) node-edge-joint related factor which is reflected by edge density and the size of node icon.}
    \label{fig:traveling_z2}
\end{figure*}

For WR graphs, we utilize the color of node icon to reflect the node-related factor $b$; and the number of neighboring rings in the graph topology to reflect the edge-related factor $a$; and the density of graph edges as well as the size of node icon to reflect the edge-node-joint related factor $c$. The values of the latent variables range in $[0,10]$ and some segment of the generated graphs to visualize, as shown in Fig.~\ref{fig:traveling_z2}. All of the proposed node-edge disentanglement models (NED-) successfully discovers and disentangle at least two of all the three types of factors, while graphVAE fails in discovering both edge-related and node-edge-joint related factors, and GDVAE fails in discovering the node-edge-joint related factors. This validates the necessities of the three types of factor disentanglement and superiority of the proposed architecture which separates the inference of node-related, edge-related and node-edge-related representations.

\subsubsection{Quantitative Evaluation}
Four quantitative evaluation metrics are tested on WS dataset on different models and compared in Table~\ref{tab:all_quantitative}. The proposed node-edge disentanglement models all shows greater superority than graphVAE and baseline GDVAE. Specifically, NED-VAE and NED-IPVAE-I both achieve $100\%$ in $\beta-M$, outperforming comparison methods by $26.9\%$ and other proposed extensions by $3.9\%$. NED-IPVAE-II achieves $97.8\%$ score in $F-M$, outperforming comparison methods by $60.3\%$ and other proposed extensions by $30.8\%$.

\begin{figure}[htb]
    \centering
    \includegraphics[width=0.47\textwidth]{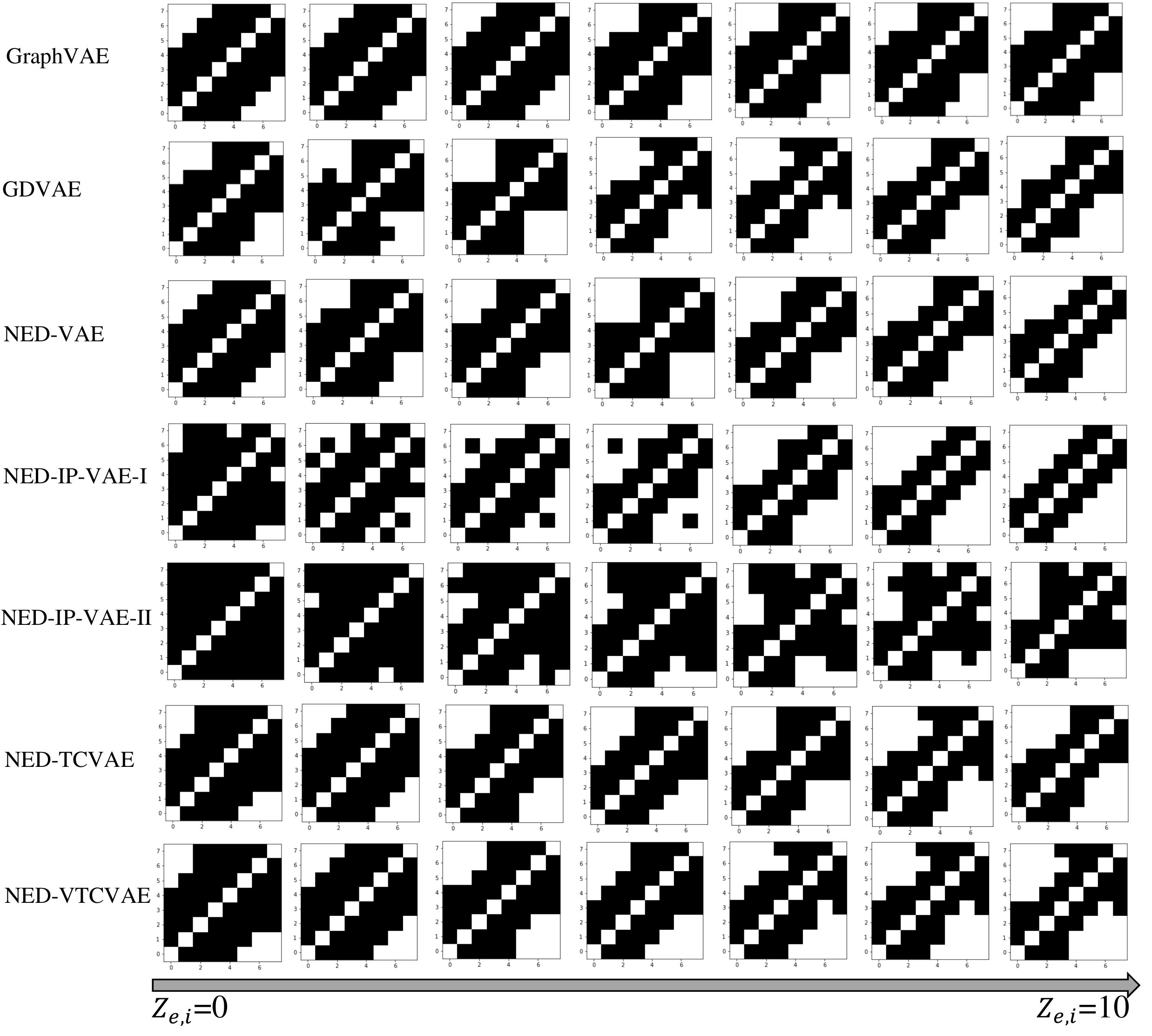}\vspace{-0.3cm}
    \caption{\small Generated contact maps from different models when one edge-related latent variable ranges from 0 to 10 in protein dataset: more blank spaces indicates higher degree of protein folding\vspace{-0.3cm}}
    \label{fig:traveling_z3}
\end{figure}

\subsection{Results for Protein Structure dataset}
\subsubsection{Qualitative Evaluation}
We evaluate the control of the factor of simulation time (T) to the generation of edges by visualizing the contact map of the proteins. The value of the relevant latent variables ranges in $[0,10]$ and some segment of the generated contact maps are shown in Fig.~\ref{fig:traveling_z3}. All of the proposed models are capable of finding T factor, while graphVAE shows bad performance in a very slight variation of structure. In addition, qualitative evaluation on protein dataset is also meaningful in analyzing how the proteins folds (reflected in contact maps) as the time flies.

\subsubsection{Quantitative Evaluation}
Four quantitative evaluation metrics are also tested on protein dataset on different models and compared in Table~\ref{tab:all_quantitative}. The proposed node-edge disentanglement models, especiallt NED-VAE all shows greater superiority than graphVAE and baseline GDVAE. Specifically, NED-VAE outperforms the comparison methods by $14.9\%$, $32.2\%$, and $13.1\%$ on metrics of $\beta-M$, DCI and Modularity respectively; and outperforms other proposed extensions by $6.8\%$, $17.2\%$, and $2.8\%$ on metrics of $\beta-M$, DCI and Modularity respectively.This proves that the proposed NED-VAE still have superiority even when there is only edge-related factor.

\section{Conclusion}
We have introduced NED-VAE, a novel and the first method for disentangling on attributed graphs as far as we know. Moreover, we propose a generic framework of objectives including various derived disentanglement penalties to solve different issues in dealing with graph structured data, such as group-wise and variable-wise disentanglement; multiple trade-off issues between reconstructed edges and nodes, and edge-related, node-related, and node-edge-joint related latent. Finally, we have performed an experimental evaluation of disentangling qualitatively and quantitatively for the proposed NED-VAE and its extensions. The comparison with graphVAE and a baseline model validates the effectiveness of the graph disentanglement architecture and the necessities of separately learning three types of latent representations.
  
\section*{Acknowledgments}
This work was supported by the National Science Foundation (NSF)
Grant No. 1755850, No. 1841520, No. 1907805, No. 1763233, a Jeffress Memorial Trust Award, NVIDIA GPU Grant, and Design Knowledge Company (subcontract number: 10827.002.120.04). 
Y. Ye's work was partially supported by the NSF Grants IIS-2027127, IIS-1951504, CNS-1940859, CNS-1946327, CNS-1814825, OAC-1940855, and the NIJ 2018-75-CX-0032.
This material is additionally based upon work supported by (while serving at) the NSF. Any opinion, findings, and conclusions or recommendations expressed in this material are those of the author(s) and do not necessarily reflect the views of the National Science Foundation.

\tiny
\bibliographystyle{ACM-Reference-Format}
\bibliography{main.bib}

\end{document}